\documentclass{article}
\usepackage[export]{adjustbox}
\usepackage{arxiv}

\usepackage{amsmath,amsfonts,bm}

\def\eqref#1{equation~\ref{#1}}

\def\1{\bm{1}}

\DeclareMathAlphabet{\mathsfit}{\encodingdefault}{\sfdefault}{m}{sl}
\SetMathAlphabet{\mathsfit}{bold}{\encodingdefault}{\sfdefault}{bx}{n}

\usepackage{xcolor}
\definecolor{lightblue}{rgb}{0.6, 0.8, 0.9}
\definecolor{darkblue}{HTML}{0c58ac}
\definecolor{darkgreen}{rgb}{0, 0.55, 0.12}
\definecolor{darkred}{rgb}{0.6,0,0}
\definecolor{yellowbg}{RGB}{255,249,224}
\definecolor{greenbg}{RGB}{232,246,232}
\definecolor{bluebg}{RGB}{226,241,255}
\definecolor{clawcolor}{RGB}{173, 43, 41}
\usepackage{natbib}
\usepackage[
  breaklinks=true,
  colorlinks,
  linkcolor=darkred,
  citecolor=clawcolor,
  bookmarks=false
]{hyperref}
\usepackage{url}
\usepackage{fontawesome}
\usepackage{footmisc}
\usepackage{tikz}
\usepackage{graphicx}
\usepackage{wrapfig}
\usepackage{booktabs}
\usepackage{adjustbox}
\usepackage{multirow}
\usepackage{xspace}
\usepackage{array}
\usepackage{bm}
\usepackage{bbm}
\usepackage{xcolor}
\usepackage{amsthm, amsmath, amssymb}
\usepackage{cleveref}
\usepackage{algorithm}
\usepackage{algorithmic}

\usepackage{listings}
\usepackage{wrapfig}
\usepackage{caption}
\usepackage{subcaption}
\usepackage{url}
\usepackage{colortbl}
\usepackage{adjustbox}
\usepackage{makecell}
\usepackage{pifont}
\usepackage{tcolorbox}
\usepackage{setspace}
\usepackage{fancybox}
\usepackage{tocloft}
\usepackage{etoc}
\usepackage{enumitem}
\usepackage{pifont}
\usepackage{bm}
\usepackage{ulem}
\usepackage[table]{xcolor}

\definecolor{darksilver}{RGB}{160,160,160} 
\definecolor{mymauve}{rgb}{0.58,0,0.82}
\newcommand{\inlineicon}[1]{\raisebox{-0.15em}{\includegraphics[height=1em]{#1}}}

\newlength{\mysize}

\allowdisplaybreaks

\theoremstyle{definition}

\title{{\includegraphics[width=2cm, valign=m]{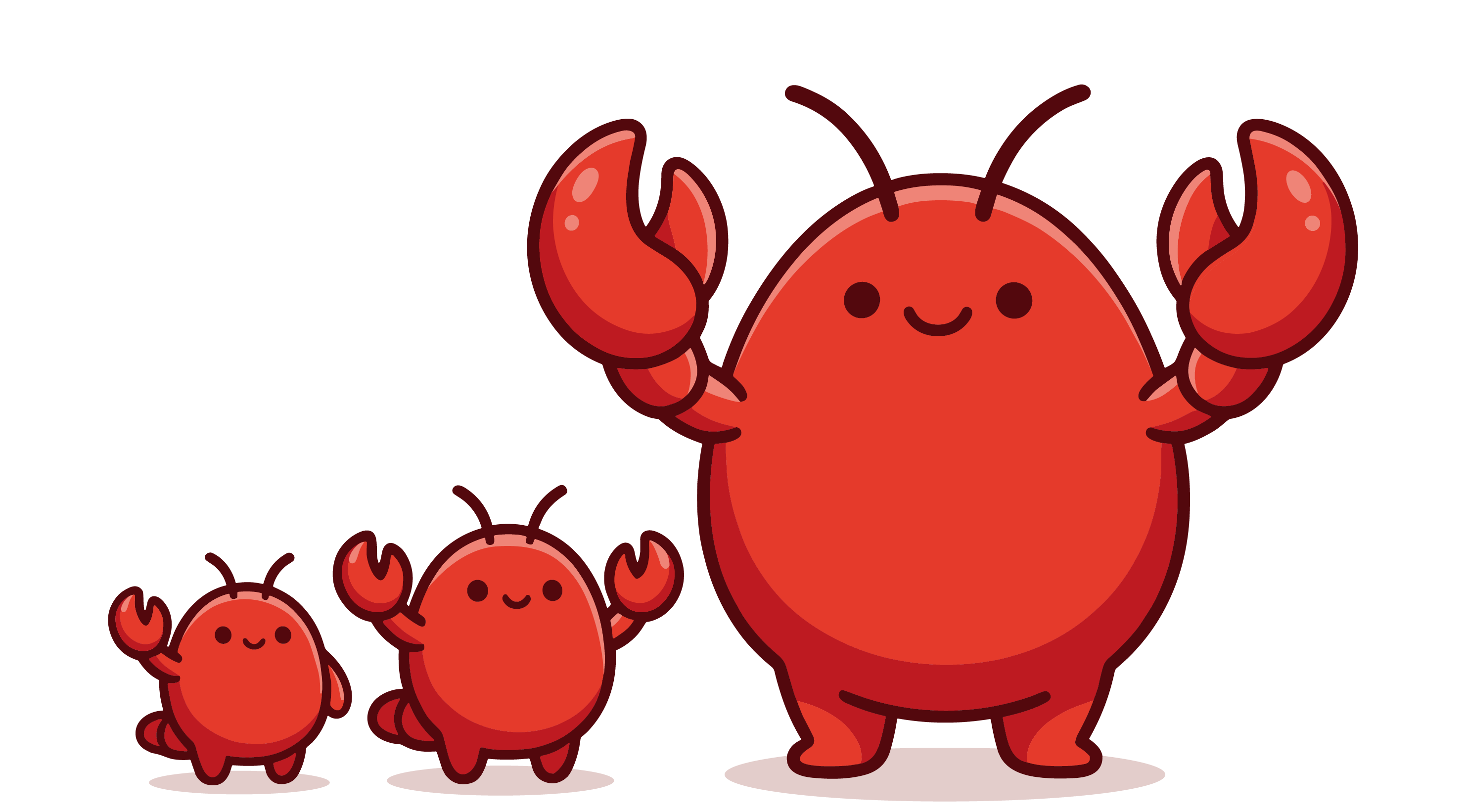}} QuantClaw: Precision Where It Matters for OpenClaw}
\author{
Manyi Zhang$^{1}$ \quad 
Ji-Fu Li$^{1}$\thanks{Corresponding author.} \quad
Zhongao Sun$^{1}$ \quad
Xiaohao Liu$^{2}$\\
\textbf{Zhenhua Dong}$^{1}$\quad 
\textbf{Xianzhi Yu}$^{1}$\quad
\textbf{Haoli Bai}$^{1}$\quad
\textbf{Xiaobo Xia}$^{3}$\\
$^1$Huawei Technologies\quad
$^2$National University of Singapore \\
$^3$University of Science and Technology of China\\
\small\tt \{zhangmanyi6@huawei.com~~lijifu4@huawei.com~~xiaoboxia@ustc.edu.cn\}
}
\begin{document}

\maketitle

\begin{tcolorbox}[
    colback=gray!8,     
    colframe=gray!0,     
    boxrule=0pt,        
    left=2mm, right=2mm,  
    top=5mm, bottom=5mm,  
    coltitle=black,    
    arc=2mm
  ]
\vspace{-0.35cm}
\begin{abstract}
Autonomous agent systems such as OpenClaw introduce significant efficiency challenges due to long-context inputs and multi-turn reasoning. This results in prohibitively high computational and monetary costs in real-world development. While quantization is a standard approach for reducing cost and latency, its impact on agent performance in realistic scenarios remains unclear. In this work, we analyze quantization sensitivity across diverse complex workflows over OpenClaw, and show that precision requirements are highly task-dependent. Based on this observation, we propose QuantClaw, a plug-and-play precision routing plugin that dynamically assigns precision according to task characteristics. QuantClaw routes lightweight tasks to lower-cost configurations while preserving higher precision for demanding workloads, saving cost and accelerating inference without increasing user complexity. Experiments show that our QuantClaw maintains or improves task performance while reducing both latency and computational cost. Across a range of agent tasks, it achieves up to 21.4\% cost savings and 15.7\% latency reduction on GLM-5 (FP8 baseline). These results highlight the benefit of treating precision as a dynamic resource in agent systems.
\end{abstract}
\begin{quote}
\vskip 0.2in
\inlineicon{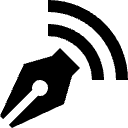} Blog: \href{https://sparkengineai.github.io/QuantClaw}{https://sparkengineai.github.io/QuantClaw}\\
\inlineicon{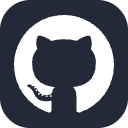} GitHub:  \href{https://github.com/SparkEngineAI/QuantClaw-plugin}{https://github.com/SparkEngineAI/QuantClaw-plugin}\\
\inlineicon{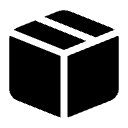} ClawHub:  \href{https://clawhub.ai/plugins/@sparkengineai%2Fquantclaw}{https://clawhub.ai/plugins/@sparkengineai/quantclaw}
\end{quote}
\end{tcolorbox}
\addtocontents{toc}{\protect\setcounter{tocdepth}{-1}}

\section{Introduction}
\label{sec:intro}
Recent advances in large language models have enabled the emergence of autonomous agents capable of executing complex and multi-step workflows in realistic environments~\cite{wu2024autogen,li2023camel,luo2025gui,lu2026ui,zhao2026agentic,huang2024understanding,ferrag2025llm,putta2024agent,dong2025survey,liu2023dynamic,jin2025stella,gao2024empowering}. Systems such as OpenClaw~\cite{openclaw2026openclaw}, Hermes~\cite{nousresearch2026hermesagent}, and Claude Code~\cite{anthropic2025claudecode} extend language models beyond static text generation by equipping them with tool use, environment interaction, and iterative reasoning capabilities. These systems function as general-purpose agent frameworks, where a model is responsible not only for producing outputs, but for planning, acting, and coordinating actions across heterogeneous tasks.

However, this shift toward agentic systems introduces new computational challenges. Unlike traditional inference workloads, agent execution often involves long context accumulation, tool output storage, and multi-step interactions, resulting in significantly increased latency and cost~\cite{zhang2024chain,xiao2025improving,ding2026calibrate,sui2025stop}, which quickly becomes prohibitively expensive in practical deployments. For instance, a single user session may accumulate over 234K tokens of context with OpenClaw~\cite{apiyi2026openclaw}. Consequently, even a routine follow-up query necessitates transmitting the full historical state to the model, greatly increasing the cost per interaction. In current practice, these systems typically operate at a fixed precision or model configuration, regardless of task complexity. This results in a systematic mismatch between 
resource allocation and actual task requirements, making OpenClaw-style systems inherently cost-inefficient.

Quantization~\cite{liu2024llm, li2026batquant,liu2026freeact,li2024svdquant,liu2024spinquant, sun2024flatquant} offers a promising direction for improving efficiency by reducing memory footprint and accelerating inference. However, its impact on agent performance remains poorly understood, especially in complex multi-turn collaboration scenarios~\cite{dong2025can}. In particular, while quantization has been extensively studied in standard natural language processing benchmarks~\cite{liu2025quantization, huang2024empirical, zhao2025benchmarking, zhang2026benchmarking}, its effect in real-world agentic scenarios, where tasks vary widely in reasoning depth, tool use, and interaction patterns, has not been systematically characterized.

In this work, we investigate how quantization affects different categories of agent tasks within OpenClaw. We first construct a benchmark-driven analysis~\cite{ye2026claw} (including 24 distinct task types, 104 tasks, and 6 models at scales ranging from 9B to 744B) to evaluate quantization sensitivity across diverse workloads, revealing that the impact of reduced precision is highly task-dependent. Based on these findings, we derive a set of practical insights on when and where high precision is necessary. Building on these insights, we propose QuantClaw, a plug-and-play precision routing plugin for OpenClaw. QuantClaw dynamically assigns precision levels according to task characteristics, routing lightweight tasks to lower-cost configurations while preserving higher precision for more demanding workloads. This design enables more cost-effective and faster services without introducing additional complexity for end users.

Empirically, QuantClaw achieves superior performance-efficiency trade-offs through task-adaptive precision routing.
On GLM-4.7-Flash (PinchBench v1.2.0\footnote{\url{https://github.com/pinchbench}}), QuantClaw improves the average score by 2.85 points over the BF16 baseline while simultaneously reducing cost by 21.6\% and latency by 8.4\%. On the larger GLM-5 (PinchBench v2.0.0), it achieves an average score gain of 2.09 points over FP8, with 21.4\% cost savings and 15.7\% latency reduction. These results suggest that precision should be treated as a dynamic resource in agent systems, rather than a fixed configuration.

\vspace{-12.5pt}
\section{Empirical Findings and Insights}\label{sec:findings}
\vspace{-5pt}
\subsection{Setups}
\textbf{Evaluation benchmark.} We utilize Claw-Eval~(release v0.0.0)~\cite{ye2026claw} as our primary evaluation benchmark, which is an end-to-end evaluation suite designed for autonomous agents. It comprises 24 task types and 104 human-verified tasks spanning multiple domains, including service orchestration, multimodal perception, and multi-turn dialogue, and evaluates agents along completion, safety, and robustness dimensions. Importantly, Claw-Eval incorporates trajectory-level auditing and controlled perturbation, enabling a more reliable assessment of agent behavior beyond final outputs. 

\textbf{Models and measurement metrics.} We employ 6 models for benchmarking, which include GLM-4.7-Flash-30B\footnote{\url{https://huggingface.co/zai-org/GLM-4.7-Flash}}~\cite{zeng2025glm}, GLM-5-744B\footnote{\url{https://huggingface.co/zai-org/GLM-5}}~
\cite{zeng2026glm}, MiniMax-M2.5-229B\footnote{\url{https://huggingface.co/MiniMaxAI/MiniMax-M2.5}}, Qwen3.5-9B\footnote{All models are sourced from the official Hugging Face collections: \url{https://huggingface.co/collections/Qwen/qwen35}\label{fn:qwen}}, Qwen3.5-35B-A3B\footref{fn:qwen}, and Qwen3.5-397B-A17B\footref{fn:qwen}. These models are widely adopted and representative of current large language model families, which cover a diverse range of model sizes and architectures. Note that GLM-5~\cite{zeng2026glm} is deployed in FP8 precision\footnote{While GLM-5 is not an FP8-native model, FP8 serves as its default precision setting. Therefore, we evaluate GLM-5 under FP8 in our experiments.}, while the remaining models are provided in BF16 by default, enabling a consistent comparison of quantization effects across different precision settings.

We further quantize these models from their native high-precision formats to lower-precision regimes, down to NVFP4~\cite{alvarez2025introducing, chen2025int, cook2025four,zhao2026unleashing}, enabling a systematic investigation of precision-performance trade-offs. Our analysis is conducted at both coarse-grained and fine-grained levels, capturing overall performance degradation as well as task-dependent sensitivity across different categories of agent workloads.

We evaluate both native (with BF16 or FP8 precision) and quantized models (with NVFP4 precision) along three key dimensions: \textit{Score}, \textit{Cost}, and \textit{Speed}. \textit{Score} reflects task performance, measuring the agent’s ability to complete tasks under different precision settings successfully. \textit{Cost} quantifies computational expense, including memory usage and inference-related compute. \textit{Speed} captures efficiency, typically measured in latency or throughput during agent execution. This metric framework enables a systematic comparison between full-precision and low-precision models, allowing us to analyze precision-performance trade-offs at both the overall and task-dependent levels. Besides, each experimental case was executed over 6 independent trials to reduce the randomness of the results.

\subsection{Global Quantization Effects}
\textbf{Main results.} To understand the overall impact of quantization on OpenClaw-based agent systems, we evaluate the mentioned models under both native precision (BF16/FP8) and low-precision NVFP4 settings. The results are summarized in Table~\ref{tab:quantization_scaling}. Since cost gains from quantization vary across hardware platforms, inference frameworks, and model architectures, following~\cite{nvidia2025moe, spheron2026fp4, knoop2026private}, we estimate NVFP4 token pricing at 80\% of the BF16 price, where the BF16 baseline is referenced from the official OpenRouter pricing\footnote{\url{https://openrouter.ai/}}. Overall, quantization to NVFP4 introduces only limited performance degradation across models, indicating that aggressive precision reduction can be effectively applied in agent workloads. However, the impact is not uniform and exhibits a clear dependence on model scale. For smaller models (\textit{e.g.}, Qwen3.5-9B), we observe noticeable performance drops (approximately 3-4\%), suggesting that low-capacity models are more sensitive to precision reduction. In contrast, medium-scale models (\textit{e.g.}, 30B-70B range) show moderate degradation, typically within 2\%. More interestingly, large-scale models (200B+) demonstrate strong robustness to quantization, with performance drops below 2\%, and in some cases even slight improvements. For example, GLM-5 and MiniMax-M2.5 exhibit small performance gains after quantization, indicating that reduced precision may introduce a regularization effect in certain regimes. This trend suggests a scaling effect in quantization robustness: as model size increases, sensitivity to precision reduction decreases. Intuitively, larger models possess greater representational redundancy, making them more tolerant to quantization noise.

\begin{table}[!t]
\centering
\small
\caption{\textbf{Model performance, cost, and latency under native precision (BF16 or FP8) and NVFP4 quantization.} Large-scale models show strong robustness to quantization, and in some cases even exhibit slight performance gains.}
\resizebox{\textwidth}{!}{%
\begin{tabular}{l|c|c|ccc|ccc}
\toprule
\multirow{2}{*}{\textbf{Model}} &  \multirow{2}{*}{\textbf{Params (B)}} & \multirow{2}{*}{\textbf{Native Precision}} & \multicolumn{3}{c|}{\textbf{BF16/FP8}} & \multicolumn{3}{c}{\textbf{NVFP4}} \\
& & & \textbf{Score$\uparrow$} & \textbf{Cost (\$)$\downarrow$} & \textbf{Time (s)$\downarrow$} & \textbf{Score$\uparrow$} & \textbf{Cost (\$)$\downarrow$} & \textbf{Time (s)$\downarrow$} \\
\midrule
GLM-4.7-Flash     & 30  & BF16 &0.6370 & 0.0077 & 72.20 & 0.6034 & 0.0072 & 73.02 \\
GLM-5             & 744 & FP8 & 0.7130 & 0.0647 & 68.96 & 0.7229 & 0.0548 & 87.58 \\
MiniMax-M2.5      & 229 & BF16 & 0.6760 & 0.0112 & 44.89 & 0.6823 & 0.0084 & 59.27 \\
Qwen3.5-9B        & 9 & BF16  & 0.4267 & 0.0022 & 16.58 & 0.4107 & 0.0013 & 16.99 \\
Qwen3.5-35B-A3B   & 35 & BF16 & 0.6686 & 0.0300 & 59.61 & 0.6549 & 0.0235 & 55.49 \\
Qwen3.5-397B-A17B & 397 & BF16 & 0.7048 & 0.0539 & 62.10 & 0.6937 & 0.0441 & 42.46 \\
\bottomrule
\end{tabular}%
}
\label{tab:quantization_scaling}
\end{table}

\begin{figure}[!t]
    \centering
    \includegraphics[width=0.85\linewidth]{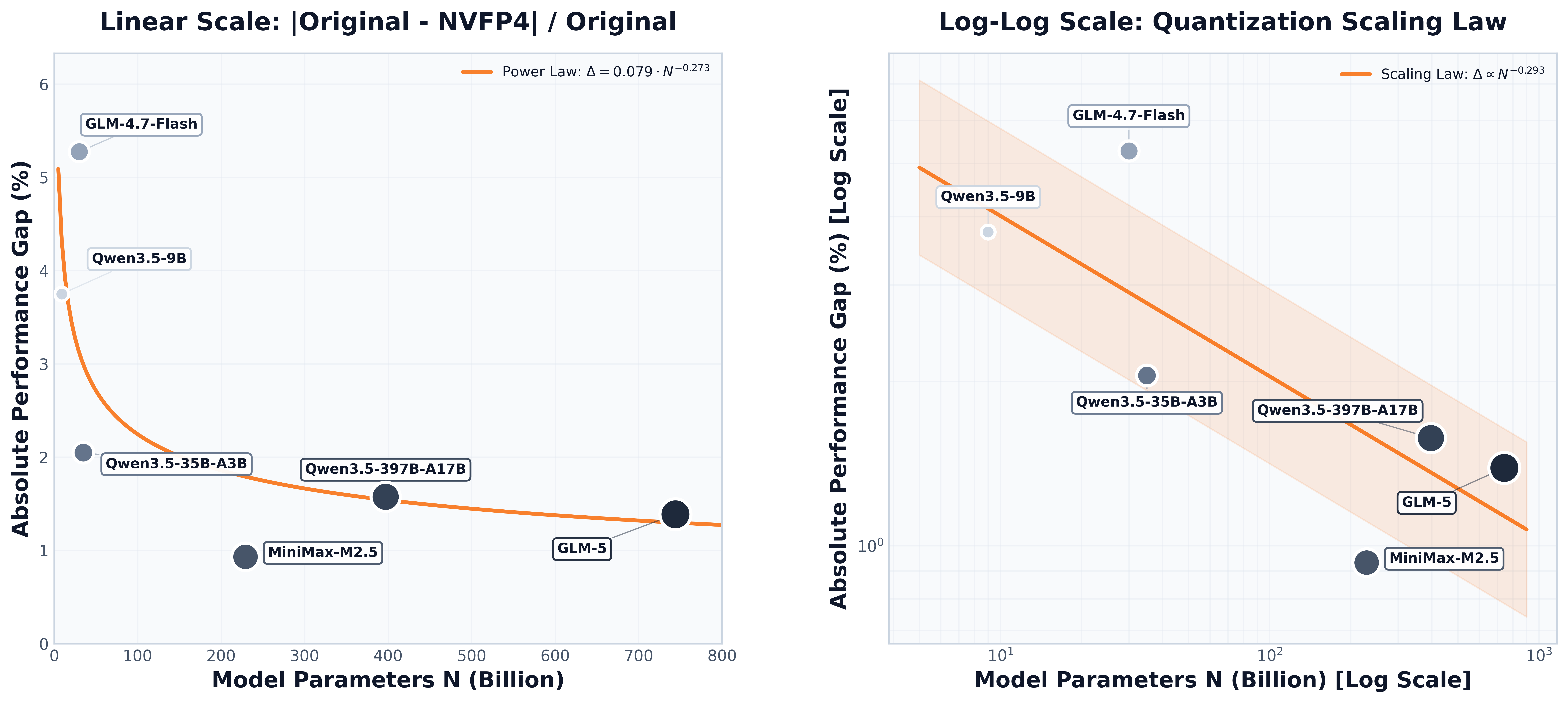}
    \caption{\textbf{Scaling behavior of quantization degradation under NVFP4.} \textit{Left}: Absolute performance gap vs. model size on a linear scale, showing diminishing degradation as model parameters increase. \textit{Right}: Log-log plot reveals a power-law relationship, confirming systematic scaling. Larger models demonstrate enhanced robustness to low-precision quantization, with reduced sensitivity compared to smaller counterparts.}
    \label{fig:scaling}
\end{figure}

\textbf{Scaling effect of quantization degradation under NVFP4.} We observe a consistent scaling behavior of quantization degradation with model size. On the linear scale (see Figure~\ref{fig:scaling}(\textit{Left})), the performance gap diminishes as model parameters increase, indicating improved robustness, with the power law $\Delta=0.079\cdot N^{-0.273}$. On the log-log scale, this trend follows a power-law $\Delta\propto N^{-0.293}$ (see Figure~\ref{fig:scaling}(\textit{Right})), revealing that the reduced sensitivity is governed by a systematic scaling law rather than incidental variation. This scaling behavior implies that aggressive low-precision deployment becomes increasingly viable as model size grows. In practice, large-scale models can tolerate substantial precision reduction with minimal performance loss, 
suggesting that precision optimization should be prioritized for smaller or mid-sized models where sensitivity is higher.

\begin{figure}[!t]
    \centering
    \includegraphics[width=0.9\linewidth]{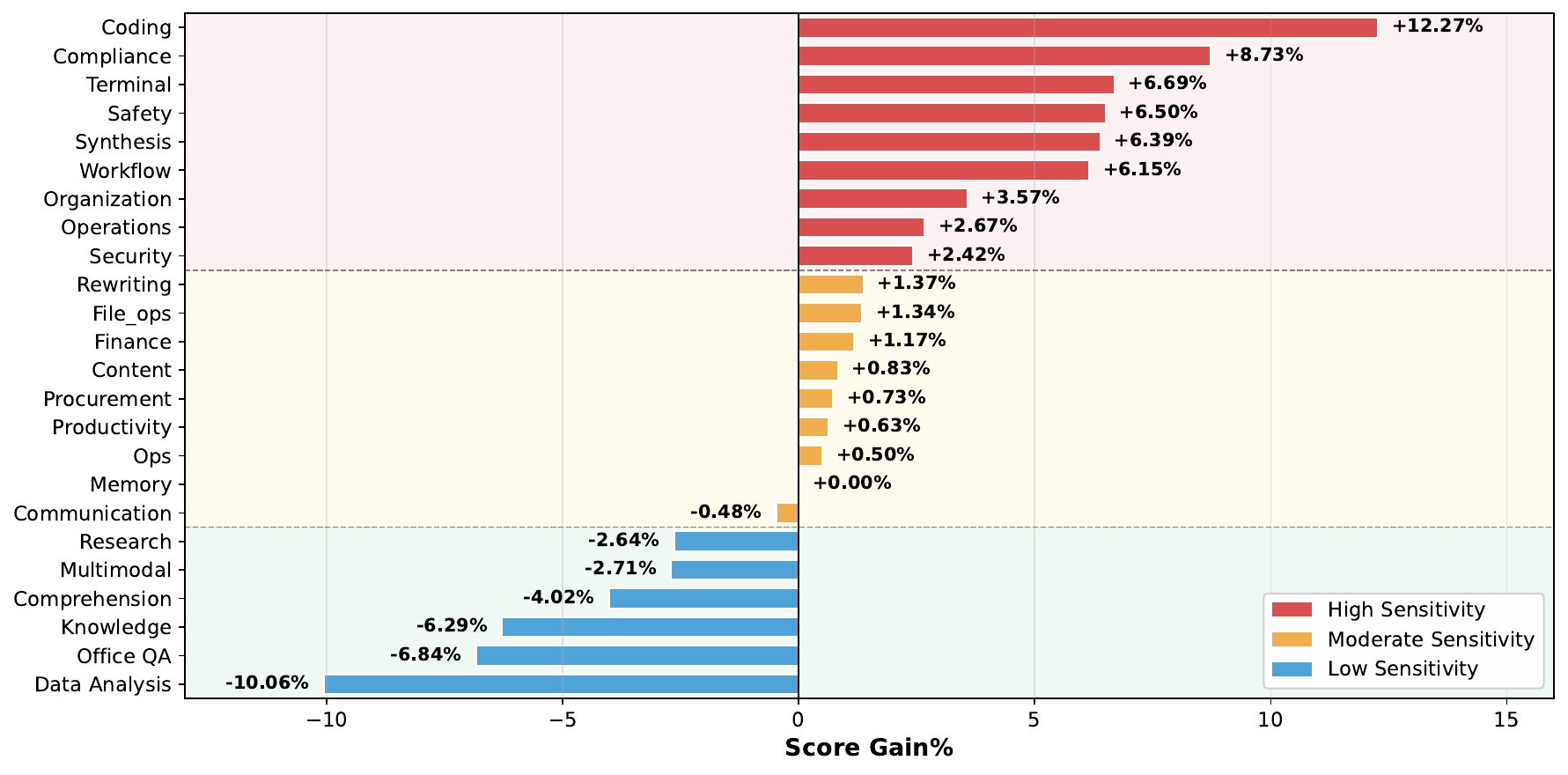}
    \caption{\textbf{Task-level variability in quantization sensitivity.} Positive values indicate performance drops after quantization from BF16/FP8 to NVFP4, while negative values indicate relative performance gains, \textit{e.g.}, (BF16 score - NVFP4 score)/(BF16 score). Tasks are grouped into high-, moderate-, and low-sensitivity categories according to their observed response to quantization.}
     \label{fig:task_sensitivity_all}
\end{figure}

\subsection{Task-Level Quantization Sensitivity}\label{sec:task_sen}
The impact of low-precision quantization varies significantly across task types. Averaging the results across all test models, we categorize OpenClaw tasks into three groups based on sensitivity: high, moderate, and low (Figure~\ref{fig:task_sensitivity_all}). High-sensitivity tasks (\textit{e.g.}, code, compliance, terminal, safety-critical) suffer notable degradation under NVFP4 due to their reliance on precise decision boundaries. Low-sensitivity tasks (\textit{e.g.}, research, comprehension, retrieval, analysis) remain robust and may even benefit from quantization, likely due to their tolerance to approximation. Moderate-sensitivity tasks (\textit{e.g.}, rewriting, content generation) show minimal changes and can operate under mixed precision. These observations highlight the task-dependent nature of quantization, motivating task-aware precision allocation.

\begin{figure}[t]
    \centering
    \includegraphics[width=\linewidth]{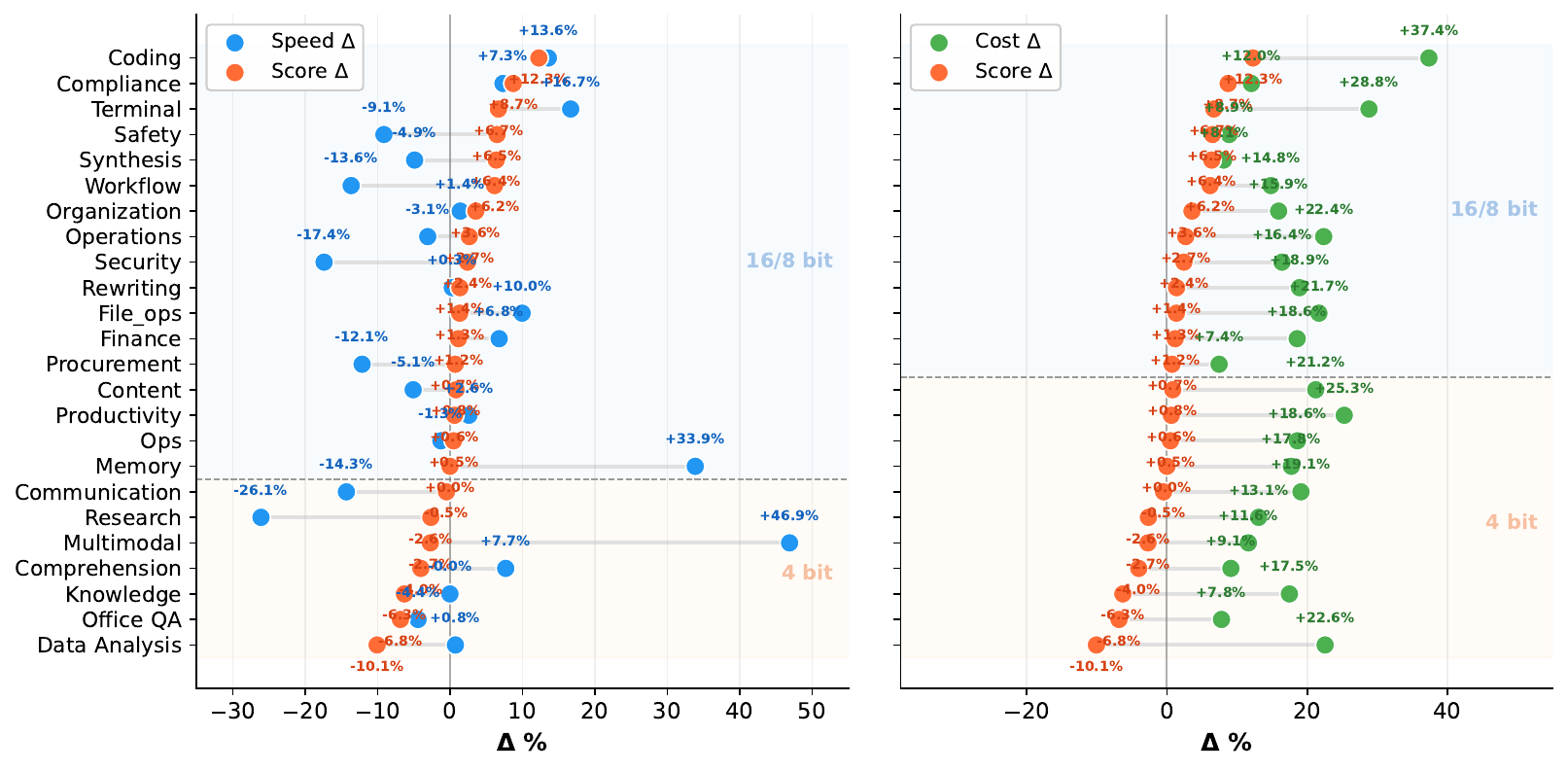}
    \caption{\textbf{Task-level deployment trade-off between 16/8-bit and 4-bit.} Score \textit{vs.} speed (\textit{left}) and score \textit{vs.} cost (\textit{right}), ranked by the sensitivities. $\Delta \%$ indicates the relative value. Tasks in the 16/8-bit zone show quality-critical precision dependence, where tasks in the 4-bit zone are cost-efficient low-precision candidates.}
    \label{fig:score_speed_cost}
\end{figure}

\subsection{Routing Views and Policy}
The findings above suggest that precision should not be assigned uniformly across OpenClaw tasks. Since different task types exhibit markedly different sensitivity to quantization, QuantClaw introduces a task-aware routing policy that selects the execution precision according to the expected trade-off among score, speed, and cost. In practice, we propose two modes that enable different users for their usage:
\begin{itemize}
    \item Latency-oriented: tasks are routed to lower-precision execution when the reduction in latency outweighs the negligible loss in performance
    \item Cost-oriented: lower precision is preferred for tasks whose quality remains stable under quantization, thereby reducing computational expense
\end{itemize}

The score-speed and score-cost comparisons and their preferred task-wise precisions are summarized in Figure~\ref{fig:score_speed_cost}.

\section{Methodology}
\subsection{Motivation}
The empirical analysis in Section~\ref{sec:findings} suggests that quantization affects OpenClaw tasks unevenly. While overall degradation from BF16/FP8 to NVFP4 is limited, the impact varies substantially across task types. Tasks such as code generation, compliance, and safety-critical decision making appear more sensitive to precision reduction, whereas research, comprehension, and analytical tasks remain robust and may even benefit from low precision. These observations are treated as indicative trends that motivate system design rather than definitive conclusions. They suggest that a uniform precision policy is suboptimal, motivating adaptive precision allocation based on task characteristics.

QuantClaw addresses this by treating precision as a runtime-controllable resource. It operates at the service-provider level, adapting execution precision for a fixed model without requiring user intervention. As a result, users do not need to manage trade-offs between quality, cost, and latency, which are handled transparently by the system.

\begin{figure}
    \centering
    \includegraphics[width=\linewidth]{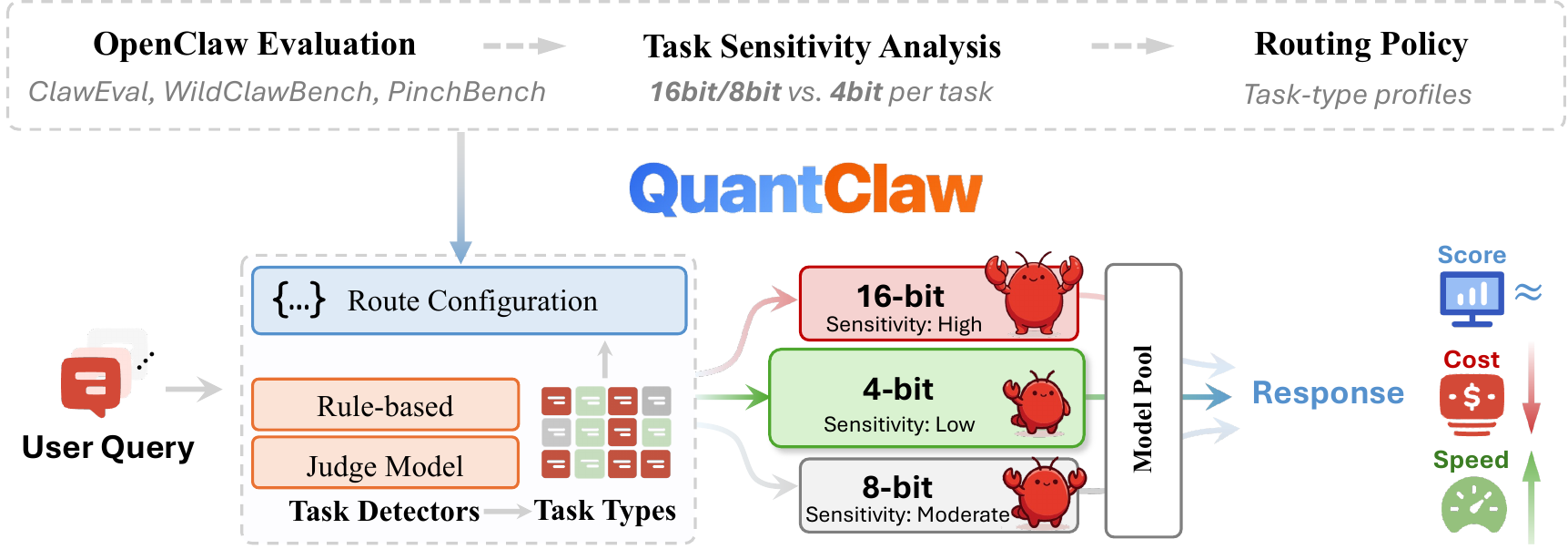}
    \caption{\textbf{Overall pipeline of QuantClaw.} The user query is first identified as different task categories, supported by a hybrid detection mechanism. QuantClaw then performs precision routing based on precomputed task–precision sensitivity profiles. A pool of model variants is maintained to support preferred precision for deployment.}
    \label{fig:overfiew}
\end{figure}

\subsection{QuantClaw Pipeline}
The overview of the QuantClaw pipeline is offered in Figure~\ref{fig:overfiew}. We detail it below. 

\textbf{Task detection and routing interface.} Given a user query, QuantClaw first performs task identification to determine the task category, which serves as the primary signal for subsequent routing decisions. This is implemented through a hybrid detection mechanism: rule-based detectors use predefined patterns, keywords, and simple structural cues (\textit{e.g.}, format or interaction patterns) for explicit cases. For queries that are not captured by rule-based detectors, model-based detectors take over and resolve ambiguous cases via a lightweight classifier. The design is intentionally modular. These detectors are interchangeable and can be replaced or extended with more advanced methods, since the core objective of this stage is simply to produce a reliable task-type label rather than depend on a specific detection strategy.

\textbf{Precision routing mechanism.} After task identification, QuantClaw performs precision routing based on precomputed task–precision sensitivity profiles (see Section~\ref{sec:task_sen}). The system maintains a pool of model variants at different precision levels (\textit{e.g.}, 16-bit, 8-bit, and 4-bit), and each task type is mapped to a preferred precision derived from offline analysis. At runtime, routing is straightforward: high-sensitivity tasks are executed with high precision to preserve reliability, low-sensitivity tasks are assigned to low precision to maximize efficiency, and intermediate cases are resolved according to deployment objectives (\textit{e.g.}, latency or cost constraints). This design enables consistent gains without per-query optimization overhead.

\subsection{QuantClaw System Properties}

\begin{figure}[t]
    \centering
    \begin{minipage}[t]{0.48\linewidth}
        \centering
\includegraphics[width=\textwidth]{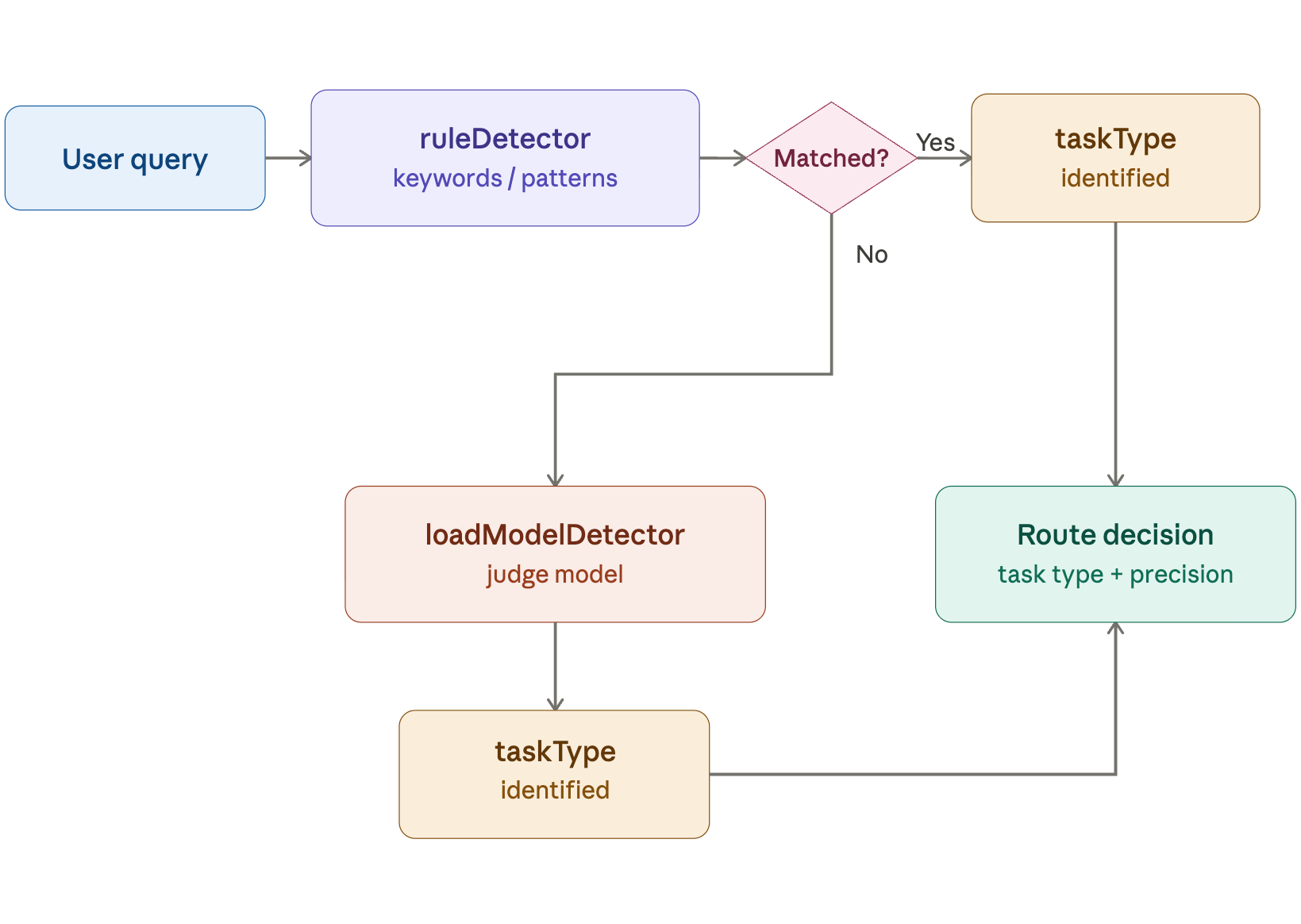}
        \caption{\textbf{Automatic Adaptation} that consolidates both task detectors.}
        \label{fig:training1}
    \end{minipage}\hfill
    \begin{minipage}[t]{0.48\linewidth}
        \centering
\includegraphics[width=\textwidth, clip, trim=12cm 0 0 0.2cm]{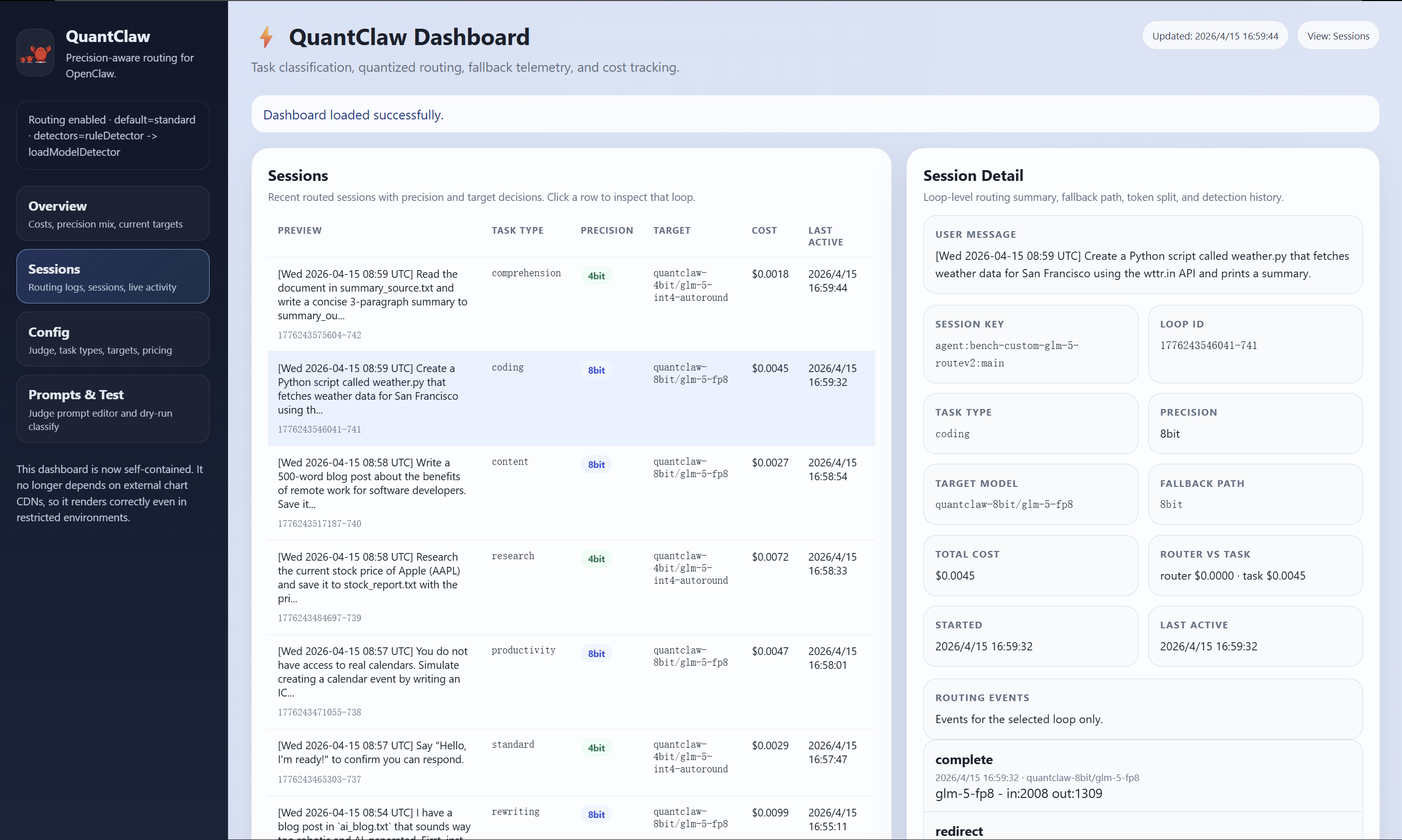}
        \caption{\textbf{Intelligent Routing} supported to make decisions on-the-fly.}
        \label{fig:training2}
    \end{minipage}\hfill
    \begin{minipage}[t]{0.48\linewidth}
        \centering
\includegraphics[width=\textwidth, clip, trim=12cm 2.3cm 0 0]{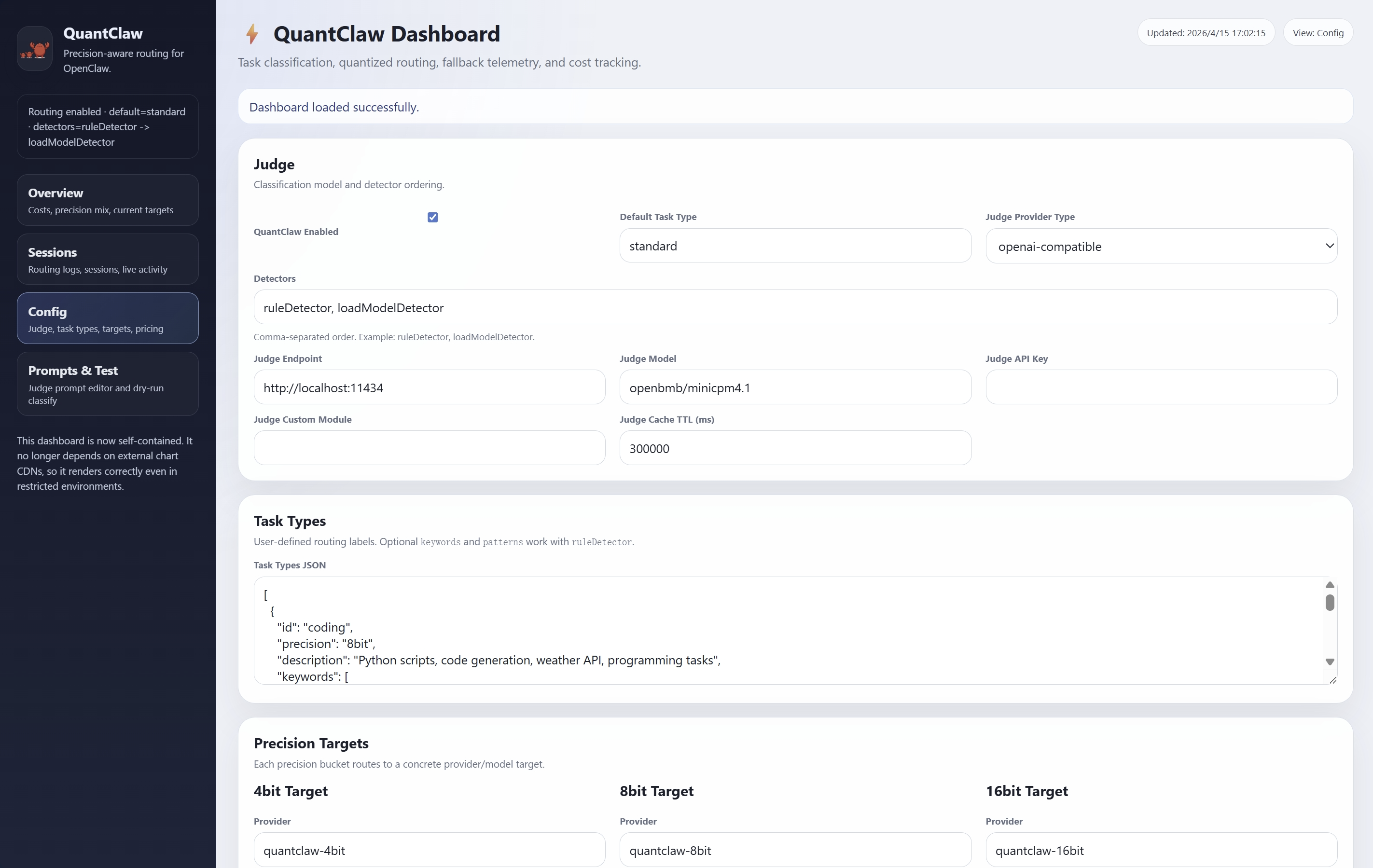}
        \caption{\textbf{Full Customizability} for interaction with dashboard interface.}
        \label{fig:training3}
    \end{minipage}\hfill
    \begin{minipage}[t]{0.48\linewidth}
        \centering
\includegraphics[width=\textwidth, clip, trim=5.8cm 0 0 0]{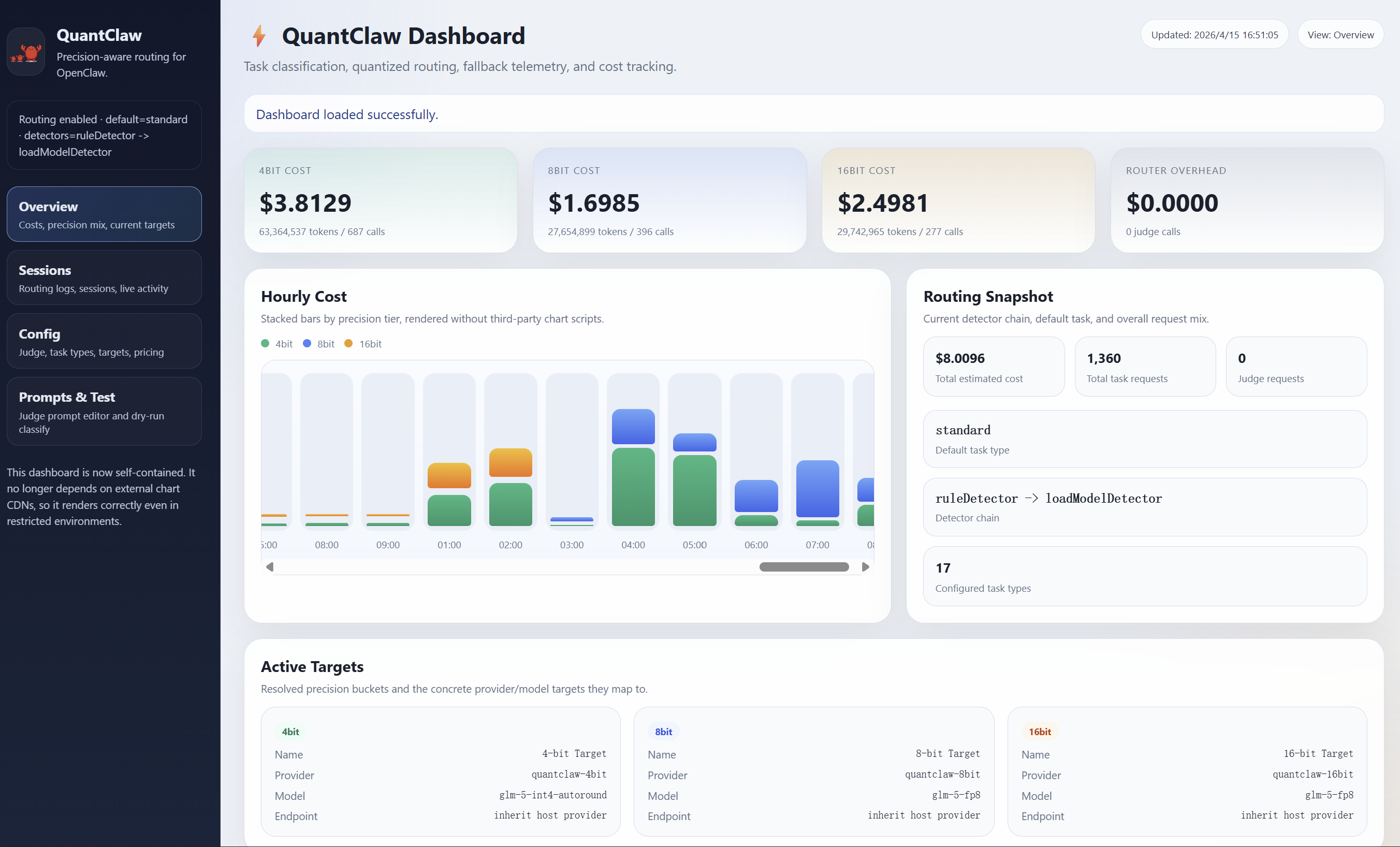}
        \caption{\textbf{Built-in Dashboard and Observability} on overview status.}
        \label{fig:training4}
    \end{minipage}

\end{figure}

QuantClaw is designed as a practical runtime layer with several key system-level properties. It performs automatic adaptation by making routing decisions on the fly based on inferred task characteristics, without requiring any user involvement. The system is lightweight to deploy, operating as a plug-in layer on top of existing models. Finally, QuantClaw provides built-in observability, exposing routing decisions along with cost and performance metrics in real time, enabling transparent monitoring and fine-grained control in production settings. We provide the related diagrams in Figure~\ref{fig:training1}, Figure~\ref{fig:training2}, Figure~\ref{fig:training3}, and Figure~\ref{fig:training4}.

\section{Benefits of QuantClaw}

\begin{table}[!tp]
\centering
\caption{\textbf{Comparison of QuantClaw with fixed-precision baselines on PinchBench.} QuantClaw achieves better performance-efficiency trade-offs, reducing cost and latency while maintaining or improving task performance.} 
\begin{tabular}{l|l|l|c|c|c}
\toprule
\textbf{Benchmark} & \textbf{Model} & \textbf{Method} & \textbf{Score (Best / Avg)$\uparrow$} & \textbf{Cost (USD)$\downarrow$} & \textbf{Time (s)$\downarrow$} \\
\midrule

\multirow{6}{*}{\textbf{PinchBench v1.2.0}} & \multirow{3}{*}{GLM-4.7-Flash}
& All BF16   & 81.57 / 81.26 & 0.001598 & 19.07 \\
& & All INT4   & 82.63 / 78.71 & 0.001422 & 21.80 \\
& & \cellcolor{gray!15}QuantClaw  & \cellcolor{gray!15}85.46 / \textbf{84.11} & \cellcolor{gray!15}\textbf{0.001252} & \cellcolor{gray!15}\textbf{17.47} \\

\cmidrule{2-6}
& \multirow{3}{*}{GLM-5}
& All FP8    & 87.65 / 87.08 & 0.0127 & 34.53 \\
& & All INT4   & 90.10 / 88.24 & 0.0105 & 32.19 \\
& & \cellcolor{gray!15}QuantClaw  & \cellcolor{gray!15}90.09 / \textbf{89.09} & \cellcolor{gray!15}0.0119 & \cellcolor{gray!15}33.21 \\

\midrule

\multirow{6}{*}{\textbf{PinchBench v2.0.0}} & \multirow{3}{*}{GLM-4.7-Flash}
& All BF16   & 78.19 / 76.95 & 0.00233 & 57.10 \\
& & All INT4   & 75.24 / 73.87 & 0.00232 & 54.60 \\
& & \cellcolor{gray!15}QuantClaw  &  \cellcolor{gray!15}79.78 / \textbf{76.95} & \cellcolor{gray!15}\textbf{0.00228} &  \cellcolor{gray!15}\textbf{52.35} \\

\cmidrule{2-6} 
& \multirow{3}{*}{GLM-5}
& All FP8    & 85.72 / 83.50  & 0.0196 & 62.22 \\
& & All INT4   &  89.31 / 81.92 & 0.0169 & 58.99 \\
& & \cellcolor{gray!15}QuantClaw  & \cellcolor{gray!15}87.25 / \textbf{85.59} & \cellcolor{gray!15}\textbf{0.0154} & \cellcolor{gray!15}\textbf{52.46} \\

\bottomrule
\end{tabular}
\label{tab:quantclaw_results}
\end{table}

\begin{table}[!tp]
  \centering
  \caption{\textbf{Comparison of different task detection methods}. QuantClaw supports rule- and judge model-based methods for task detection. `+' indicates a hybrid strategy for task detection. 
  }
  \label{tab:model_performance}
    \begin{tabular}{l|c|c|c}
    \toprule
    \textbf{Method} & \textbf{Accuracy (\%)$\uparrow$} & \textbf{Macro F1 (\%)$\uparrow$} & \textbf{Avg. Time (s/query)$\downarrow$} \\
    \midrule
    RuleDetector                & 83.13 & 65.90 & 0.0017 \\
    \midrule
    BGE-M3                      & 89.76 & 86.56 & 0.0200 \\
    \midrule
    GLM-4.7-Flash-INT4                & 86.75 & 83.68 & 0.0800 \\
    GLM-5-FP8                   & 92.17 & 89.72 & 0.1717 \\
    MiniCPM-4.1                 & 87.35 & 82.25 & 1.5627 \\
    \midrule
    RuleDetector + BGE-M3       & 91.53 & 88.66 & 0.0149 \\
    RuleDetector + GLM-5-FP8    & 93.37 & 91.30 & 0.1217 \\
    RuleDetector + MiniCPM-4.1  & 90.05 & 87.35 & 0.5900 \\
    \bottomrule
  \end{tabular}
\end{table}

\textbf{Setups.} We conduct evaluations on PinchBench~(v1.2.0 and v2.0.0) here. This benchmark includes diverse OpenClaw-style workloads, making it a suitable testbed for observing how different task types respond to precision variation. We evaluate two model backbones, GLM-4.7-Flash and GLM-5, under three numerical settings: a higher-precision baseline (BF16 for GLM-4.7-Flash and FP8 for GLM-5), a uniform low-precision configuration (INT4), and QuantClaw, which adaptively selects precision based on task characteristics, enabling a direct comparison between static precision schemes and dynamic allocation. The measurement metrics include \textit{Score}, \textit{Cost}, and \textit{Time}. Note that we select PinchBench instead of ClawEval and adopt INT4 quantization rather than NVFP4, as described in Section~\ref{sec:findings}. This is because we aim to verify the \textit{generalizability} of our findings about task-level variability in Section~\ref{sec:task_sen}.

\textbf{Results and trade-off analysis.} Based on the throughput comparison in Table~\ref{tab:bf16_int4_throughput}~(see Appendix~\ref{sec:A}), INT4 delivers an average throughput gain of 14.34\% over BF16 under the same latency constraints. Considering pricing simplicity and industry discounting practice~\cite{kurtic2025give}, we set the INT4 token price to 85\% of the BF16 price. Table~\ref{tab:quantclaw_results} presents the end-to-end evaluation of QuantClaw against fixed-precision baselines on PinchBench v1.2.0 and v2.0.0. We report best and average task scores, per-sample inference cost (USD), and end-to-end latency (seconds). As can be seen, QuantClaw consistently outperforms fixed-precision baselines by achieving a better score-efficiency trade-off. On GLM-4.7-Flash (PinchBench v1.2.0), QuantClaw improves average performance to 84.11 (vs. 81.26 BF16 and 78.71 INT4), while reducing latency to 17.47s (8.4\% latency improvement) and lowering cost to 0.001252 (21.7\% reduction). On GLM-5, QuantClaw maintains competitive performance (89.09 vs. 87.08 FP8 and 88.24 INT4), while slightly reducing latency (33.21s vs. 34.53s, 1.04$\times$ speedup) and cost (0.0119 vs. 0.0127, -6.3\%). The benefits are more ponounced on v2.0.0 for GLM-5. QuantClaw improves the average score by 2.09 point over FP8 baseline, while achieving 21.4\% cost savings and 15.7\% speed up.

The ablation study on task detection methods is shown in Table~\ref{tab:model_performance}. QuantClaw supports various detection methods, including individual detectors such as RuleDetector, an embedding model (\textit{i.e.}, BGE-M3~\cite{chen2024bge}), a model-as-judge (\textit{e.g.}, GLM-4.7-Flash-INT4), and a hybrid strategy. Introducing judge models improves the detection accuracy while increasing the time overhead. However, the hybrid strategy achieves an acceptable trade-off, demonstrating the highest accuracy and a reasonable time cost. This establishes \textit{RuleDetector + BGE-M3} as the default choice for QuantClaw.

These results indicate that the gain of QuantClaw comes not from uniformly lowering precision, but from selectively applying low precision to tolerant tasks while preserving high precision for sensitive ones, yielding a strictly better operating point than either high-precision or uniform low-precision execution.

\section{Conclusion}
\label{sec:conclusion}
We show that uniform precision is not only inefficient but economically impractical for OpenClaw systems, as quantization sensitivity varies significantly across tasks. To address this, we propose QuantClaw, a task-aware precision routing mechanism that dynamically allocates precision based on workload characteristics. QuantClaw improves efficiency by reducing latency and cost while maintaining or even improving performance. These results suggest that precision should be treated as a dynamic and task-dependent resource, providing a simple yet effective direction for optimizing agent systems.

\section{Outlook}
Current personal AI systems often rely on a single model operating at uniformly high precision, regardless of task complexity. This results in unnecessary computational cost and latency, as many requests do not require maximum capability. QuantClaw suggests an alternative paradigm within OpenClaw: precision and model capability should be allocated dynamically based on task requirements. By routing lightweight tasks to lower-precision configurations and reserving stronger resources for more demanding workloads, QuantClaw improves overall system efficiency without increasing user complexity. More broadly, this points to a shift in how OpenClaw systems can evolve. Instead of simply integrating multiple models, they should actively coordinate them. In this sense, QuantClaw serves as a concrete step toward treating precision and model capability as allocatable resources in a multi-model system.

\clearpage



\bibliography{main}
\bibliographystyle{unsrt}
\clearpage
\appendix

\section{Additional Results}\label{sec:A}
\begin{table}[htbp]
\centering

\caption{Maximum concurrency and output throughput of GLM-4.7-Flash in BF16 and INT4, where at least 90\% of requests satisfy TTFT $\leq$ 500 ms and TPOT $\leq$ 10 ms. INT4 achieves a 14.34\% average throughput gain over BF16 across all evaluated settings.}

\label{tab:bf16_int4_throughput}
\small
\setlength{\tabcolsep}{8pt}
\begin{tabular}{cccccr}
\toprule
Input & Output & Bit & Concurrency & {Output Throughput (tok/s)} & {Gain} \\
\midrule
\multirow{2}{*}{2048} & \multirow{2}{*}{4096} & BF16 & 32 & 3326.00 & \multirow{2}{*}{24.01\%} \\
                      & & INT4 & 40 & 4124.73 &                         \\
\addlinespace

\multirow{2}{*}{2048} & \multirow{2}{*}{8192} & BF16 & 32 & 3078.00 & \multirow{2}{*}{7.12\%} \\
                      & & INT4 & 32 & 3297.00 &                        \\
\addlinespace

\multirow{2}{*}{4096} & \multirow{2}{*}{2048} & BF16 & 32 & 3164.00 & \multirow{2}{*}{7.27\%} \\
                      & & INT4 & 32 & 3394.00 &                        \\
\addlinespace

\multirow{2}{*}{4096} & \multirow{2}{*}{4096} & BF16 & 24 & 2519.00 & \multirow{2}{*}{29.72\%} \\
                      & & INT4 & 32 & 3267.72 &                         \\
\addlinespace

\multirow{2}{*}{8192} & \multirow{2}{*}{1024} & BF16 & 20 & 1969.00 & \multirow{2}{*}{18.78\%} \\
                      & & INT4 & 22 & 2338.70 &                         \\
\addlinespace

\multirow{2}{*}{8192} & \multirow{2}{*}{2048} & BF16 & 20 & 1981.00 & \multirow{2}{*}{6.01\%} \\
                      & & INT4 & 20 & 2100.00 &                        \\
\addlinespace

\multirow{2}{*}{8192} & \multirow{2}{*}{4096} & BF16 & 18 & 1829.77 & \multirow{2}{*}{13.86\%} \\
                      & & INT4 & 20 & 2083.31 &                         \\
\addlinespace

\multirow{2}{*}{8192} & \multirow{2}{*}{8192} & BF16 & 16 & 1680.58 & \multirow{2}{*}{7.96\%} \\
                      & & INT4 & 18 & 1814.30 &                        \\
\midrule
\multicolumn{5}{r}{Average Throughput Gain} & 14.34\% \\
\bottomrule
\end{tabular}
\end{table}

\end{document}